\documentclass[letterpaper]{article}

\usepackage[switch]{lineno}
\usepackage{xspace}
\newcommand{\acronym}{MCDC\xspace}
\newcommand{\name}{Mixing Consistent Deep Clustering\xspace}
\newcommand{\githubLink}{www.github.com}

\usepackage{aaai}
\usepackage{times}
\usepackage{helvet}
\usepackage{courier}
\usepackage{graphicx}
\usepackage{amsmath}

\usepackage{algorithm}
\usepackage{algorithmic}

\usepackage{caption}

\begin{document}

\title{\name}

%% comment out authors for double blind review
% \author{Paper ID: 
% }
\author{Daniel Lutscher \\
dLutscher@gmail.com\\
VU University Amsterdam \\
Amsterdam, The Netherlands
\And Ali el Hassouni\\
a.el.hassouni@vu.nl\\
VU University Amsterdam\\
Amsterdam, The Netherlands\\
\AND Maarten Stol\\
maarten.stol@braincreators.com\\
BrainCreators B.V.\\
Amsterdam, The Netherlands\\
\And Mark Hoogendoorn\\
m.hoogendoorn@vu.nl\\
VU University Amsterdam\\
Amsterdam, The Netherlands
}

\maketitle
\begin{abstract}
Finding well-defined clusters in data represents a fundamental challenge for many data-driven applications, and largely depends on good data representation. Drawing on literature regarding representation learning, studies suggest that one key characteristic of good latent representations is the ability to produce semantically mixed outputs when decoding linear interpolations of two latent representations. We propose the Mixing Consistent Deep Clustering (\acronym) method which encourages interpolations to appear realistic while adding the constraint that interpolations of two data points must look like one of the two inputs. By applying this training method to various clustering (non-)specific autoencoder models we found that using the proposed training method systematically changed the structure of learned representations of a model and it improved clustering performance for the tested ACAI, IDEC, and VAE models on the MNIST, SVHN, and CIFAR-10 datasets. These outcomes have practical implications for numerous real-world clustering tasks, as it shows that the proposed method can be added to existing autoencoders to further improve clustering performance.
\end{abstract}

%--------------------------------------
%	SECTION: Introduction
%--------------------------------------
\section{Introduction}\label{sec:introduction}
Clustering is the process of assigning data points to groups based on similarity. It constitutes a fundamental part of many data-driven applications and over the past decades a large variety of clustering techniques have been studied \cite{rokach2005clustering,arthur2007k,paparrizos2015k}. When applying these techniques to high-dimensional data, they are typically preceded by dimensionality reduction techniques such as Principal Component Analysis (PCA) \cite{wold1987principal} and spectral methods \cite{ng2002spectral} to overcome the often prevalent limitation of data sparsity. Through recent advances in Deep Learning \cite{lecun2015deep,schmidhuber2015deep}, deep neural networks have since also been used as dimensionality reduction techniques. Among existing approaches, autoencoders \cite{hinton2006reducing} have widely been used to achieve this task by learning latent representations in an unsupervised setting. In the context of clustering, the successfully trained encoder can then be used to transform the input data into a lower-dimensional latent representation, which in turn serves as the input for clustering techniques. This combination of Deep Learning and clustering is called \textit{Deep Clustering}. Nowadays, several novel architectures emerged from this new research field, mostly by combining an autoencoder and a clustering algorithm through simultaneous optimization \cite{guo2017improved,jiang2016variational,min2018survey}.

While the end-to-end optimization shows promising results, a limitation of this approach is that the models are solely trained towards creating non-ambiguous clusters. In other words, they maximize the distance between already existing clusters by minimizing a data point’s distance to its assigned cluster centroid. However, they do not change the learned position of latent representations in a way that it would create more \textit{disentangled} clusters. Disentanglement refers to how well the underlying factors of variation are separated \cite{bengio2013representation}. One relevant characteristic of disentanglement in the context of clustering is the occurrence of natural clusters. It is suggested that natural clusters may occur because the local variations on a manifold tend to reflect a single category, and a linear interpolation between datapoints of different classes involves going through a low density region \cite{bengio2013representation} (see Figure \ref{fig:interpolation} for a schematic illustration). Thus, while traversing the linearly interpolated path between instances of two clusters, one should not encounter instances of a third cluster. Since the positions of the lower-dimensional feature representations are learned, this interpolation characteristic can be learned as well. This training method, often referred to as \textit{interpolation training}, has been used in supervised settings to increase the robustness of a classifier \cite{verma2018manifold,guo2019mixup} as well as in a semi-supervised setting to achieve comparable performance with fewer class instances \cite{verma2019interpolation,berthelot2019mixmatch}. 

This paper makes use of this interpolation training technique in an unsupervised setting. We use a mixing function that linearly interpolates two latent representations using an $\alpha$ coefficient drawn from a uniform distribution. To ensure that the output of mixed latent representations look like stemming from the data distribution at pixel level, we make use of adversarial training \cite{goodfellow2014generative} by training a discriminator to predict the $\alpha$ coefficient used for the mixing function. Here, the goal of the autoencoder is to fool the discriminator that the outputs of mixed representations are real by always predicting an $\alpha$ coefficient of 0. Furthermore, the autoencoder's loss function includes an additional 'mixing consistency' loss term. The motivation for this term is to enforce the desideratum of linear dependencies between clusters in the latent space, which realizes a structure that is better suited for downstream clustering tasks (see \ref{sec:methods} for an illustration).

This new framework, called \acronym, can be applied to preexisting convolution-based autoencoder models to further improve clustering performance. In summary, the contributions of this work are as follows:
\begin{itemize}
    \item We combine the approaches of interpolation training and adversarial learning with the explicit enforcement of a desired structure in the latent space.
    \item Experimentally, we show that \acronym outperforms all other tested models with regard to clustering results on the MNIST, SVHN and CIFAR-10 datasets (section \ref{sec:results}).
    \item We empirically show that this improved framework can be added to various autoencoder architectures to improve clustering performance.
\end{itemize}

%--------------------------------------
%	SECTION: RELATED WORK
%--------------------------------------
\section{Related Work}\label{sec:related_work}

\subsection{Deep Clustering}
Recent work in Deep Clustering started to focus on creating a clustering-friendly feature space through the simultaneous minimization of reconstruction and clustering loss. As one of the first to combine both steps into an end-to-end framework, Guo and colleagues \shortcite{guo2017improved} published the \textit{Improved Deep Embedded Clustering} (IDEC) model, an improved version of the Deep Embedded Clustering (DEC) algorithm \cite{xie2016unsupervised}, that simultaneously learns feature representations and cluster assignments using an autoencoder. It first pretrains the autoencoder by using a reconstruction loss and then fine-tunes the network by adding a cluster assignment hardening loss, where each point is assigned a probability of belonging to a given cluster. This simultaneous optimization approach significantly increased clustering accuracy compared to separate autoencoder and clustering optimization \cite{guo2017improved}. Nowadays, the architecture of the IDEC model is typically used as a basis for clustering-specific autoencoder models and it has been extended to generative models such as variational autoencoders (VAE) \cite{kingma2013auto} and generative adversarial networks (GAN) \cite{goodfellow2014generative}. For example, Jiang et al. \shortcite{jiang2016variational} developed the \textit{variational deep embedding} (VaDE) framework, which simultaneously optimizes a Variational Autoencoder and a clustering-specific Gaussian Mixture Model (GMM). This results in a significant improvement in clustering accuracy as they outperformed the standard GMM clustering method, VAE and GMM networks with separate optimization as well as other deep clustering methods \cite{jiang2016variational}.

\subsection{Interpolation Training}\label{sec:interpolation_training}
Learning good representations is an active field of research in itself \cite{bengio2013representation,van2017neural,wang2016structural} and primarily investigates how to learn representations that are \textit{disentangled}. Since a disentangled representation unravels the underlying structure, it displays several attributes that are useful in the context of clustering, most importantly lower dimensional manifolds, naturally occurring clusters, and a simplicity of factor dependencies (for a comprehensive list of attributes, see \citeauthor{bengio2013representation} \citeyear{bengio2013representation}). One common denominator across these attributes is the ability to linearly interpolate between clusters or data points of classes while traversing low density regions. Since the representations of data points in the latent space are learned, it is possible to enforce this interpolation characteristic through training. For example, recent work \cite{verma2018manifold,beckham2019adversarial} applied an interpolation training method called Manifold Mixup in a supervised setting to learn robust features. By interpolating two data points' hidden layer activations, the learned class-representations were more discriminative, compact, and had fewer directions of variance, suggesting that the model learned more disentangled representations.

In an unsupervised setting, a recent method called \textit{Adversarially Constrained Autoencoder Interpolation} (ACAI) has been proposed \cite{berthelot2018understanding} which uses interpolations of latent representations combined with an adversarial regularizer to train an autoencoder. Without a clustering-specific loss, this training procedure created a latent space structure that is on-par with current state-of-the-art deep clustering models with regard to clustering accuracy.

%--------------------------------------
%	SECTION: Novel Model
%--------------------------------------
\section{\name (\acronym)}\label{sec:methods}
Let us consider an autoencoder model $F(\cdot)$, consisting of the encoder part $f_{\phi}(x)$ and the decoder $g_{\theta}(f_{\phi}(x))$. To train an autoencoder, we minimize the following reconstruction loss:
\begin{equation}\label{overallAutoencoderLoss}
    L_{f,d} = || x - g_{\theta}(f_{\phi}(x)) ||^2 
\end{equation}
where $\phi$ and $\theta$ denote the learned parameters of the autoencoder.
Given a pair of inputs, $\{x^{(i)}, x^{(j)}\} \in X$, we would like to encode them into lower dimensional latent representations $z^{(i)} = f_{\phi}(x^{(i)})$ and $z^{(j)} = f_{\phi}(x^{(j)})$ where $\{z^{(i)}, z^{(j)}\} \in Z$. Next, we mix them using a convex combination and a mixing coefficient $\alpha$: 
\begin{equation}\label{eq:mix_function}
z^{(i)}_{\alpha} = (1 - \alpha) \cdot f_{\phi} (x^{(i)}) + \alpha \cdot f_{\phi} (x^{(j)})
\end{equation}
\noindent for some $\alpha \in [0,1]$ where $z^{(i)}_{\alpha} \in Z_{\alpha}$. Then, we run $z^{(i)}_{\alpha}$ through the decoder $g_{\theta}(z^{(i)}_{\alpha})$. To ensure that the reconstruction $\hat{x}^{(i)}_{\alpha} = g_{\theta}(z^{(i)}_{\alpha})$ resembles realistic and non-blurry samples from the data distribution, we train a discriminator $D_{\omega}(x)$ on real and mixed sets of reconstructions to predict the $\alpha$ coefficient used for the mixing function:
\begin{equation}
    \hat{\alpha} =
\begin{cases}
  D_{\omega}(z^{(i)}_{\alpha}) & \{z^{(1)}_{\alpha}), ..., z^{(m)}_{\alpha})\} \in Z_{\alpha}\\
  D_{\omega}(z^{(i)}) & \{z^{(1)}), ..., z^{(m)})\} \in Z\\
  0 & otherwise
\end{cases}
\end{equation}
where the output $\hat{\alpha}$ is a scalar value. The discriminator $D_{\omega}$ is then trained to minimize:
\begin{equation}\label{eq:lossDiscriminator}
    L_D = \underbrace{||D_\omega (\hat{x}_\alpha) - \alpha ||^2}_\text{predict $\alpha$ coefficient} + \underbrace{||D_\omega (\gamma x + (1 - \gamma)\hat{x}||^2}_\text{improves training stability}
\end{equation}
\noindent where $\gamma$ is a scalar hyperparameter.

% $\hat{X} = \{\hat{x}^{(1)}, ... , \hat{x}^{(m)}\}$ and $\hat{X}_{\alpha} = \{\hat{x}^{(1)}_{\alpha}, ... , \hat{x}^{(m)}_{\alpha} \}$ 

To produce a latent space structure with low-density regions between clusters and the ability to linearly interpolate between them, the reconstruction $\hat{x}_{\alpha}$ of a mixed representation $z_{\alpha}$ must look like realistic data points from the same cluster. Since we do not have this information in an unsupervised setting, we enforce that the construction $\hat{x}$ of a given pair of two inputs $\{x_i, x_j\} \in X$ must look like one of the input data points as follows:
\begin{equation}\label{eq:mix_loss_target}
    \hat{x} = 
    \begin{cases}
      x_i & \alpha \in [0, 0.5]\\
      x_j & \alpha \in [0.5, 1]\\
      0 & otherwise.
    \end{cases}
\end{equation}

\noindent Consequently, we propose \textit{\name} (\acronym), where part of the autoencoder's objective is to produce realistic mixed reconstructions that look like coming from the respective data points that were used for the mixing operation.The \acronym autoencoder therefore minimizes the following loss:
\begin{equation}\label{eq:lossMCDC}
    \begin{split}
    L_{f,g} = \underbrace{|| x_i - g_{\theta}(f_{\phi}(x_i)) ||^2}_\text{reconstruction} \\
    + \underbrace{\lambda ||d_\omega (\hat{x}_\alpha) ||^2}_\text{fool D with mixed reconstructions} \\
    + \underbrace{|| x_i - g_{\theta}((1 - \alpha) f_{\phi} (x_i) + \alpha f_{\phi} (x_j)) ||^2}_\text{create mixing consistent reconstructions}
    \end{split}
\end{equation}
\noindent where $\lambda$ is a scalar hyperparameter, $\alpha \in [0, 0.5]$, and the \textit{mixing consistency loss} is defined as:
\begin{equation}\label{eq:mixing_consistency_loss}
    L_{mix} = || x_k - g_{\theta}(\alpha f_{\phi} (x_i) + (1 - \alpha) f_{\phi} (x_j)) ||^2.
\end{equation}

%In equations \ref{eq:lossMCDC} and \ref{eq:mixing_consistency_loss}, $x_k$ is defined as

% ----------------------------------------------------------
% -- Algorithm Box
% ----------------------------------------------------------
% Fill out algorithm box with our MCDC optimization approach
% - Get 2 data points, run through encoder
% - Mix the latent representations
% - Take all 3 latent representations and run through decoder
% - Take all reconstructions and send to discriminator
% - Discriminator encodes all 3 reconstructions
% - Update autoencoder network
% - Including additional constraint
% - Update discriminator network
% - For Clustering: encode and use latent representation for downstream clustering method (e.g. k-means clustering algorithm)
\begin{algorithm}[ht]
\caption{ Stochastic gradient descent training of MCDC.}
\begin{algorithmic}
    \FORALL{number of training iterations}
        \FORALL{\textit{k} steps}
            \STATE \textbullet \xspace Sample batch of \textit{m} input data points $\{x^{(1)}, ..., x^{(m)}\}$ from the dataset.
            \STATE \textbullet \xspace Run the batch of \textit{m} samples through the encoder $f_{\phi}(x)$ to obtain the set of \textit{m} latent representations $Z_k = \{z^{(1)},...,z^{(m)}\}$.
            \STATE \textbullet \xspace create a copy of the set $Z_k$ and reverse the indices such that we receive the set $Z_{\hat{k}} =\{z^{(m)},...,z^{(1)}\}$. 
            \STATE \textbullet \xspace Mix the two sets $Z_k$ and $Z_{\hat{k}}$ using linear interpolation and a uniformly drawn $\alpha$-coefficient such that, for the two latent representations $z^{(1)} \in Z_k, z^{(m)} \in Z_{\hat{k}}$ the resulting mixing function is: $z^{(1)}_{\alpha} = (1 - \alpha) \cdot z^{(1)} + \alpha \cdot z^{(m)}$. This results in a new set of latent representations $Z_{\alpha} = \{z^{(1)}_{\alpha}, ..., z^{(m)}_{\alpha}\}$.
            \STATE \textbullet \xspace Run the two sets of latent representations $Z_k$ and $Z_{\alpha}$ through the decoder $g_{\theta}(z)$ to obtain two sets of reconstructions $\hat{X}_k = \{\hat{x}^{(1)}, ... , \hat{x}^{(m)}$ and $\hat{X}_{\alpha} = \{\hat{x}^{(1)}_{\alpha}, ... , \hat{x}^{(m)}_{\alpha} \}$.
            \STATE \textbullet \xspace Run the two reconstruction sets $\hat{X}_k$ and $\hat{X}_{\alpha}$ through the discriminator $D_{\omega}(\hat{x})$.
            \STATE \textbullet \xspace Update the discriminator by descending its stochastic gradient based on its loss function in Eq (\ref{eq:lossDiscriminator}).
            % \begin{equation}
            %     \Delta_{\omega} \frac{1}{m} \sum_{i=1}^{m} ||D(\hat{x}^{(i)}_\alpha) - \alpha ||^2 + ||D(\gamma x^{(i)} + (1 - \gamma)\hat{x}^{(i)}||^2.
            % \end{equation}
            \STATE \textbullet \xspace Update the autoencoder by descending its stochastic gradient based on its loss function defined in Eq (\ref{eq:lossMCDC}).
            % \begin{equation}
            %     \begin{split}
            %       \Delta_{\theta, \phi} \frac{1}{m} \sum_{i=1}^{m} || x^{(i)} - \hat{x}^{(i)} ||^2
            %       + \lambda ||D_\omega (\hat{x}_\alpha) ||^2 \\
            %       + || x^{(i)} - x^{(i)}_{\alpha} ||^2  
            %     \end{split}
            % \end{equation}
        \ENDFOR
    \ENDFOR 
    % \STATE For subsequent clustering: encode the input $x^{(i)}$ to receive $z^{(i)}$ which can then be used as input for clustering methods. In our experiment we used k-means clustering.
\end{algorithmic}
\end{algorithm}

\begin{figure}%[h]
\centering
\includegraphics[scale=0.6]{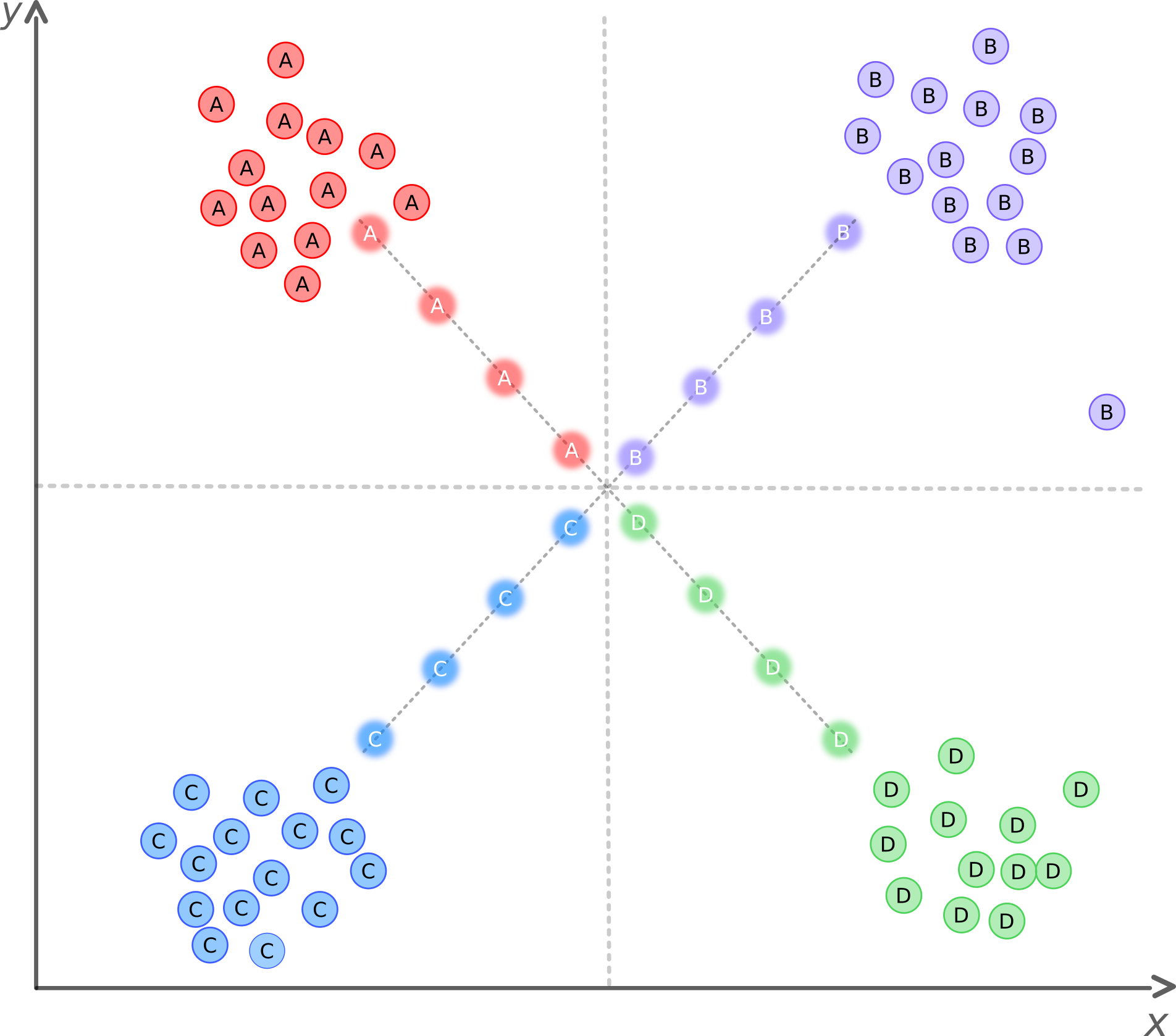}%width=6cm,height=5cm
\caption[Example Interpolation]{Illustration of linear interpolations and low-density regions between four clusters of classes.} 
\label{fig:interpolation}
\end{figure}

%% Novel Contribution / Limitation
To illustrate \acronym in the context of clustering, we first discuss an example of linear interpolations between four clusters without a mixing consistency loss, depicted in figure \ref{fig:interpolation}. Given two clusters $B=\{x^{(1)}_B, ..., x^{(i)}_B \}$ and $C=\{x^{(1)}_C, ..., x^{(j)}_C \}$, mixing the latent representations of two data points from these clusters:
\begin{equation}\label{eq:illustration_mix}
    z^{(j)}_{\alpha} = (1 - \alpha) \cdot f_{\phi}(x^{(j)}_C) + \alpha \cdot f_{\phi}(x^{(i)}_B) 
\end{equation}
results in mixed reconstructions that resemble realistic data points of cluster C for $\alpha \approx 0$ and become more semantically similar to data points from cluster B while traversing the space between the clusters as a function of increasing $\alpha$. In figure \ref{fig:interpolation}, these mixed latent representations are visualized as blurred versions of the cluster's datapoints.

Let us now consider the outlier $x^{(j)}_B \in B$, at the right end of the coordinate system. If we would mix this data point with $x^{(i)}_C \in C$ to obtain $z^{(j)}_{\alpha}$, then the reconstruction $\hat{x}^{(j)}_{\alpha}$ for $\alpha \approx 0.5$ will not necessarily resemble reconstructions of data points of cluster $B$ nor cluster $C$. This is because the vector $z^{(j)}_{\alpha}$ would lie in the area where most latent representations are decoded to resemble data points from cluster $D = \{x^{(1)}_{D}, ..., x^{(m)}_{D}\}$ (bottom right quadrant in figure \ref{fig:interpolation}).

Since the mixed reconstruction $\hat{x}^{(j)}_{\alpha}$ will look like a realistic data point, the discriminator $D_{\omega}(\hat{x}^{(j)}_{\alpha})$ will output $\hat{\alpha} = 0$. However, this behaviour would not explicitly enforce linear dependencies between clusters, which is an important characteristic of disentangled representations \cite{bengio2013representation}. Additionally, this also means that the network would not necessarily need to create more disentangled clusters because the described behaviour would not enforce the creation of low density regions between clusters. In the above scenario, the outlier from cluster B does not have any 'incentive' to move towards the cluster centroid but can stay at its position. Instead, we would like the data point to move towards the centroid position to enforce lower density regions between clusters.

\begin{figure*}
    \centering
    \includegraphics[scale=0.8]{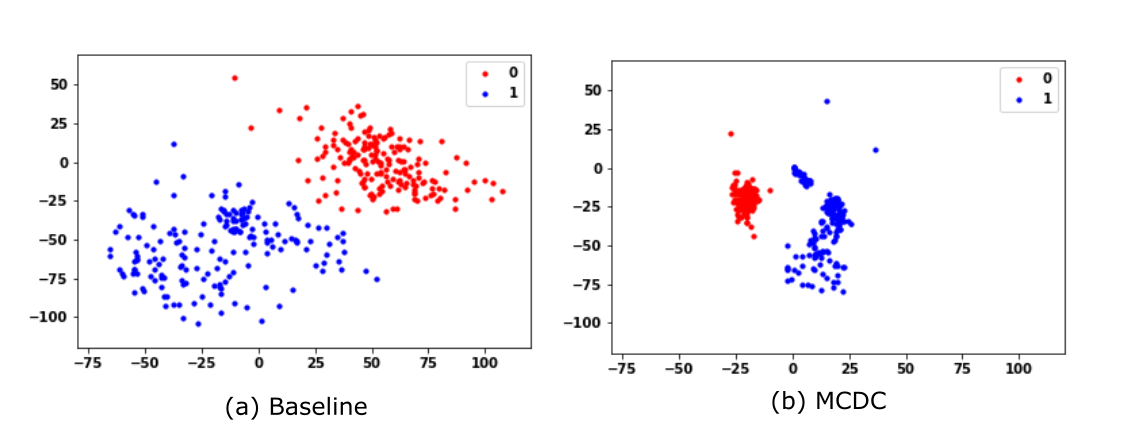}
    \caption[Example Interpolation]{The models' learned data point positions in a two-class MNIST example.}
    \label{fig:two_mnist_example}
\end{figure*}

To create this incentive of lower density regions between clusters, we introduce a mixing consistency loss to the overall autoencoder loss (eq. \ref{eq:lossMCDC}) which enforces that all mixed reconstructions look like coming from the respective data points that were used for the interpolation. Hence, in the example of mixing $x^{(j)}_B$ and $x^{(i)}_C$, where $\hat{x}^{(j)}_{\alpha}$ would resemble $\{x^{(1)}_{D}, ..., x^{(m)}_{D}\} \in D$, the overall autoencoder loss would increase as we add the mixing consistency loss (see eq: \ref{eq:mixing_consistency_loss}):
\begin{equation}
    L_{mix}=
    \begin{cases}
        || x_j - \hat{x}^{(j)}_{\alpha} ||^2 & \alpha \in [0, 0.5]\\
        || x_i - \hat{x}^{(j)}_{\alpha} ||^2 & \alpha \in [0.5, 1]\\
        0 & otherwise.
    \end{cases}
\end{equation}
Since every position in the coordinate system can only store a single representation (i.e. the vector $z_{\alpha}$ of mixed latent representations in the coordinate system can only be decoded deterministically as a data point coming from one of the clusters B, C, or D), there will be a conflict of assigning different representations to the same position. As the data point locations are learned through training, it is expected that the pressure to decode consistent representations at positions of mixed latent representations causes the outlier from cluster B to move towards the centroid position of cluster B. Eventually, this would lead to more disentangled clusters and lower-density regions between the clusters as illustrated in figure \ref{fig:two_mnist_example}.

\section{Experiments} \label{sec:experiments}
In this section, we demonstrate the effectiveness and validation of the proposed method on three benchmark datasets. We provide quantitative comparisons of \acronym with other autoencoder models. More specifically, we compare it against a convolutional VAE model to see if the \acronym framework can be combined with generative models, the IDEC model to find out if the\acronym framework can further improve on clustering-specific autoencoder models, and the ACAI framework to investigate if the added mixing consistency loss improves clustering performance. The first two models were trained twice, with and without the \acronym framework, and all models are based on the same encoder and decoder architecture. The code of \acronym is available at \githubLink.

%-----------------------------------
%	SECTION: Datasets
%-----------------------------------
\subsection{Datasets} \label{sec:method_datasets}
The following datasets were used in the experiments:

\begin{itemize}
    \item \textbf{MNIST}: The "Modified National Institute of Standards and Technology" (MNIST) dataset was first published in 1998 \cite{lecun1998mnist} and is one of the most widely used benchmark datasets. It consists of 70,000 greyscale, handwritten digits representing 10 classes. The images are centered and of size 28 by 28 pixels. To keep the same input dimensions for all datasets, the we resized the images to 32 x 32 pixels using bilinear interpolation.
    
    \item \textbf{SVHN}: The "Street View House Numbers" (SVHN) dataset consists of a total of 99,289 colored images. The images depict house numbers cropped from Google Street View. They represent 10 different classes and are of size 32 by 32 pixels \cite{netzer2011reading}.
    
    \item \textbf{CIFAR10}: The "Canadian Institute For Advanced Research" (CIFAR) developed two datasets of different sizes. For the present paper, we used the CIFAR-10 dataset consisting of 60,000 colored images of 32 x 32 pixels and 10 classes representing various objects including frogs, planes, and cars \cite{krizhevsky2009learning}.
\end{itemize}

\subsection{Experimental Setup}\label{sec:method_models}
\subsubsection{General Autoencoder Architecture}
% For an overview of the general autoencoder architecture, see Figure \ref{fig:overall_arch} in Appendix A.
% (see Appendix A, Figure \ref{fig:conv_blocks})
The overall structure of the encoder includes blocks of two consecutive 3 x 3 convolutional layers where the number of channels is doubled between the first and second layer of a given block. All convolutions are zero-padded to keep equal input and output shapes. Also, every block, except the last one, is followed by 2 x 2 average pooling. The final layer is used as the latent representation and therefore does not use an activation function while all previous layers use a leaky ReLU nonlinearity \cite{maas2013rectifier} with a negative slope of $a = 0.2$. The above-described blocks are repeated 3 times, resulting in a latent dimensionality of 256.

The decoder is created symmetrically to the encoder with respect to layer architecture, however, the average pooling function is substituted by 2 x 2 nearest neighbour upsampling \cite{odena2016deconvolution}. All parameters are initialized as zero-mean Gaussian random variables with the standard deviation:
\begin{equation}
    std = \sqrt{\frac{2}{(1 + a^2) * \text{fan\_in}}}
\end{equation}
where $a$ is the negative slope of the rectifier and fan\_in represents the number of incoming neurons from the previous layer. All models are trained with a batch size of 64 and parameters are optimized using Adam \cite{kingma2014adam} with a learning rate of 0.0001 and default values for $\beta_1$, $\beta_2$, and $\epsilon$. Across all datasets, every model is trained for a total duration of 400 epochs.

% AE architectures overview
\begin{table}
  \centering
  \caption{Overview of the different autoencoder models}
  \label{table:aeOverview}
  \begin{tabular}{c c c}
    \hline
    Model & AE Loss & Clustering Loss\\
    \hline%\midrule
        VAE & ELBO & - \\
        IDEC & MSE & KLD on SCA \\
        ACAI & MSE  & - \\
        \acronym & MSE & - \\
        \acronym+IDEC & MSE & KLD on SCA \\
        \acronym+VAE & ELBO & - \\
  \hline%\bottomrule
\multicolumn{3}{l}{\textit{Notes.} SCA=Soft Cluster Assignment}\\
\end{tabular}
\end{table}

\subsubsection{Baseline Autoencoder (AE)}
The baseline AE is the simplest model of the studied autoencoders and is used as a baseline that other models are compared against. The encoder as well as the decoder use the layer architecture and training procedure described above. In our experiments, the network tries to minimize the mean-squared error loss between the input $x$ and reconstruction $\hat{x}$. 

\subsubsection{Improved Deep Embedded Clustering (IDEC)}
The IDEC architecture \cite{guo2017improved} and pre-training procedure are based on the baseline AE. During the fine-tuning phase an additional clustering loss is added to the overall loss of the autoencoder. This clustering loss, called "cluster assignment hardening loss", consists of the KL-divergence loss of the soft cluster assignments \cite{guo2017improved}.

\subsubsection{Variational Autoencoder (VAE)}
The Variational Autoencoder (VAE) \cite{kingma2013auto} imposes a probabilistic prior distribution $p(z)$ and it trains an encoder $f_\phi(x)$ to approximate the posterior distribution $p(z|x)$ while the decoder $g_\theta(z)$ learns to reconstruct $x$ by parametrizing the likelihood $p(x|z)$. It is trained using a KL-divergence (KLD) loss and the "reparametrization trick" that replaces $z \sim \mathcal{N}(\mu,\,\sigma I)$ with $\epsilon \sim \mathcal{N}(0,\, I)$ and $ z = \mu + \sigma * \epsilon$ where $\mu, \sigma \in \mathbf{R}^{d_z}$ are the predicted mean and standard deviation produced by $f_\phi(x)$.

\begin{figure*}[hbt!]
\centering
    {\includegraphics[scale=0.5]{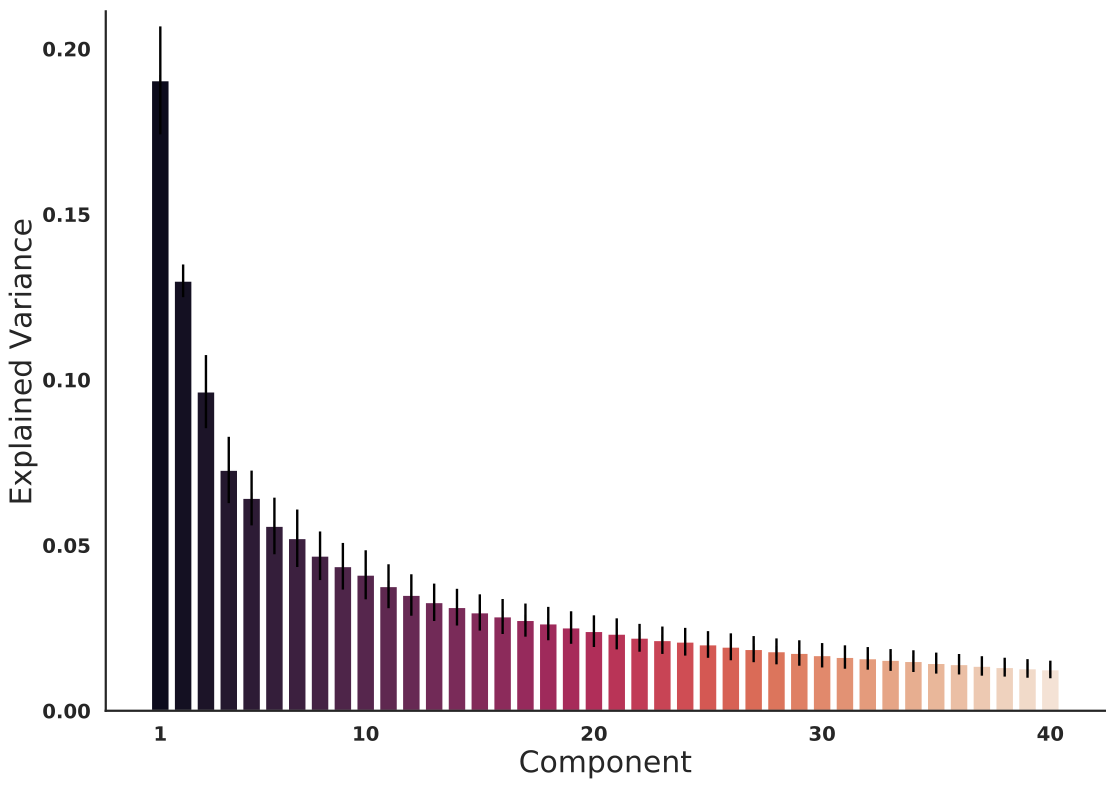}}
    \qquad
    {\includegraphics[scale=0.5]{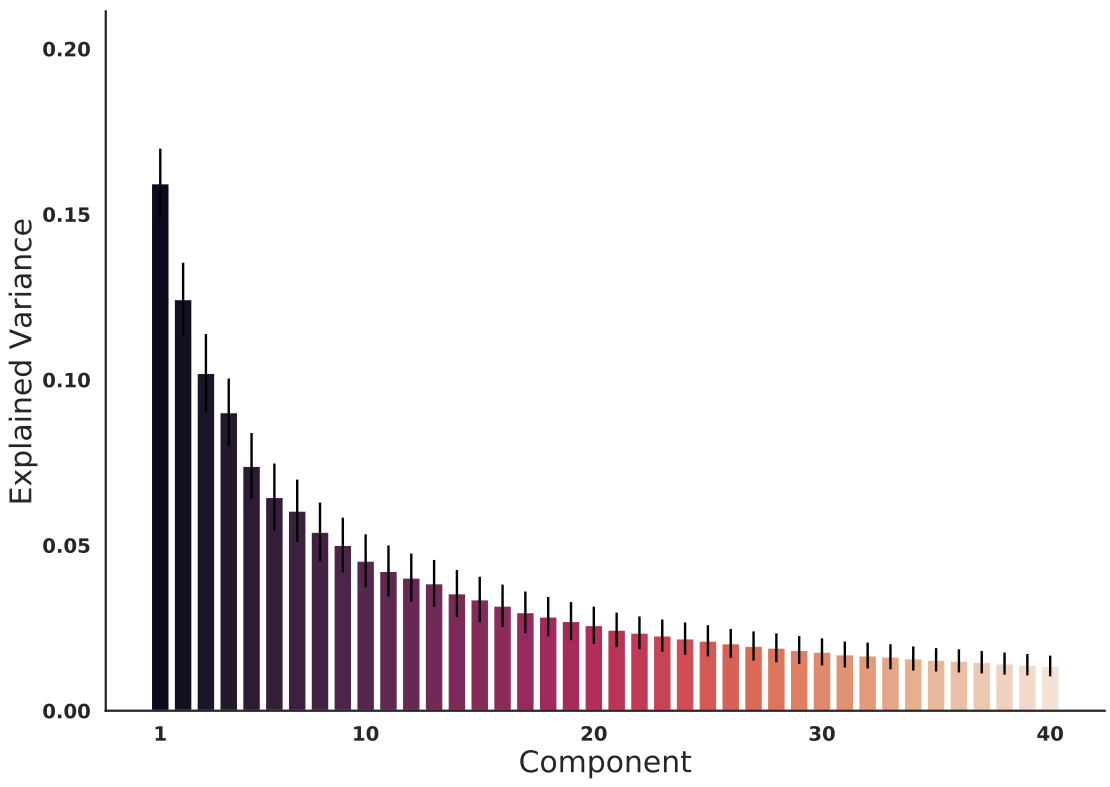}}
\caption[PCA across classes]{PCA across classes. The left panel (a) shows the PCA performance for the baseline model and the right panel (b) depicts the principal components for the \acronym model.}
\label{fig:illustration_mnist_pca}
\end{figure*}

\subsubsection{Adversarially Constrained Interpolation (ACAI)} \label{critic_section}
The ACAI model consists of two networks: the first network is a baseline autoencoder as described above. The second network is a critic that has the same layers as the baseline encoder. It receives real and interpolated reconstructions and tries to predict the $\alpha$ value used for the interpolated reconstructions. The last layer of the critic is flattened and the mean value is taken as the scalar prediction value.

\subsubsection{MCDC+IDEC and MCDC+VAE}
To address the question if the MCDC method can be combined with generative-, and clustering-specific autoencoder models, the IDEC and VAE were also trained with the MCDC method (i.e. adding the mixing function, discriminator, and mixing consistency loss), resulting in the MCDC+IDEC and MCDC+VAE models.

%-----------------------------------
%	SECTION: Evaluation Metrics
%-----------------------------------
\subsection{Evaluation Metrics} \label{sec:method_evaluation}
Clustering methods are commonly evaluated by clustering accuracy (ACC) and Normalized Mutal Information (NMI) \cite{guo2017improved,min2018survey}. We used the K-means algorithm with Euclidean distance as the distance metric. Since this algorithm is sensitive to each dimension’s relative variance, we normalized the variance prior to clustering by performing PCA whitening on the latent representations. The algorithm was run 1000 times with different random initializations and the run with the best objective value was used to calculate the accuracy and NMI of the resulting clusters. To calculate the clustering accuracy, a dataset's labels were used, as they were not used in the training process. The accuracy corresponds to the optimal one-to-one mapping of cluster IDs to classes as determined by the Hungarian algorithm \cite{kuhn2005hungarian}. This results in the following equation:
\begin{equation}\label{eq:acc}
    ACC = \underset{m}{max} \frac{\sum_{i=1}^n 1(y_i = m(c_i))}{n}
\end{equation}
where $c_i$ is the model's cluster assignment, $m$ is a mapping function that covers all possible one-to-one mappings between assignments and labels, and $y_i$ is the label. 

The second metric used in the cluster evaluation is \textit{Normalized Mutual Information (NMI)} (Estevez et al., 2009), which is a way of calculating cluster purity. The NMI calculates the mutual information score $I(Y, C)$ and then normalizes it by the amount of entropy $H$ to account for the total number of clusters. This results in the overall function:

\begin{equation}\label{eq:nmi}
    NMI(Y, C) = \frac{I(Y,C)}{\frac{1}{2}[H(Y) + H(C)]}.
\end{equation}

%-----------------------------------
%	SECTION: Results
%-----------------------------------
\section{Results}\label{sec:results}

The results section is structured as follows: we start by quantifying the intuition we presented in section \ref{sec:methods} of how adding the mixing consistency loss in \acronym affects the shape and structure of learned latent representations. Then, we show the latent embeddings learned by the models and how they performed on clustering the various datasets.

\begin{table*}[t]%[h]
    \caption{Average clustering accuracy as measured by ACC and NMI (higher is better). Clusters were calculated using 1000 random initializations of K-means on the feature space of the given autoencoder.}
    \centering
    \begin{tabular}{l c c c}
        \hline
        Model & MNIST & SVHN & CIFAR-10\\
        \hline
            ACAI & 90.08 / 88.60 & 22.00 / 15.17 & 20.30 / 09.29  \\
            IDEC & 66.03 / 72.77 & 17.97 / 05.65 & 21.75 / 09.74 \\
            VAE & 82.44 / 88.68 & 11.64 / 00.47 & 20.86 / 10.61 \\
            \acronym & 96.39 / 92.39 & \textbf{35.00} / \textbf{43.28} & \textbf{25.99} / \textbf{16.41} \\
            \acronym+IDEC & \textbf{97.61} / \textbf{93.80} & 18.36 / 09.74 & 22.17 / 10.73  \\
            \acronym+VAE & 84.95 / 89.63 & 15.24 / 01.47 & 20.85 / 09.81 \\
        \hline
    \end{tabular}
    \label{tab:clusteringResults}
\end{table*}

%% MNIST INTERPOLATIONS
\begin{figure*}%[h]
\centering
\includegraphics[scale=1.0]{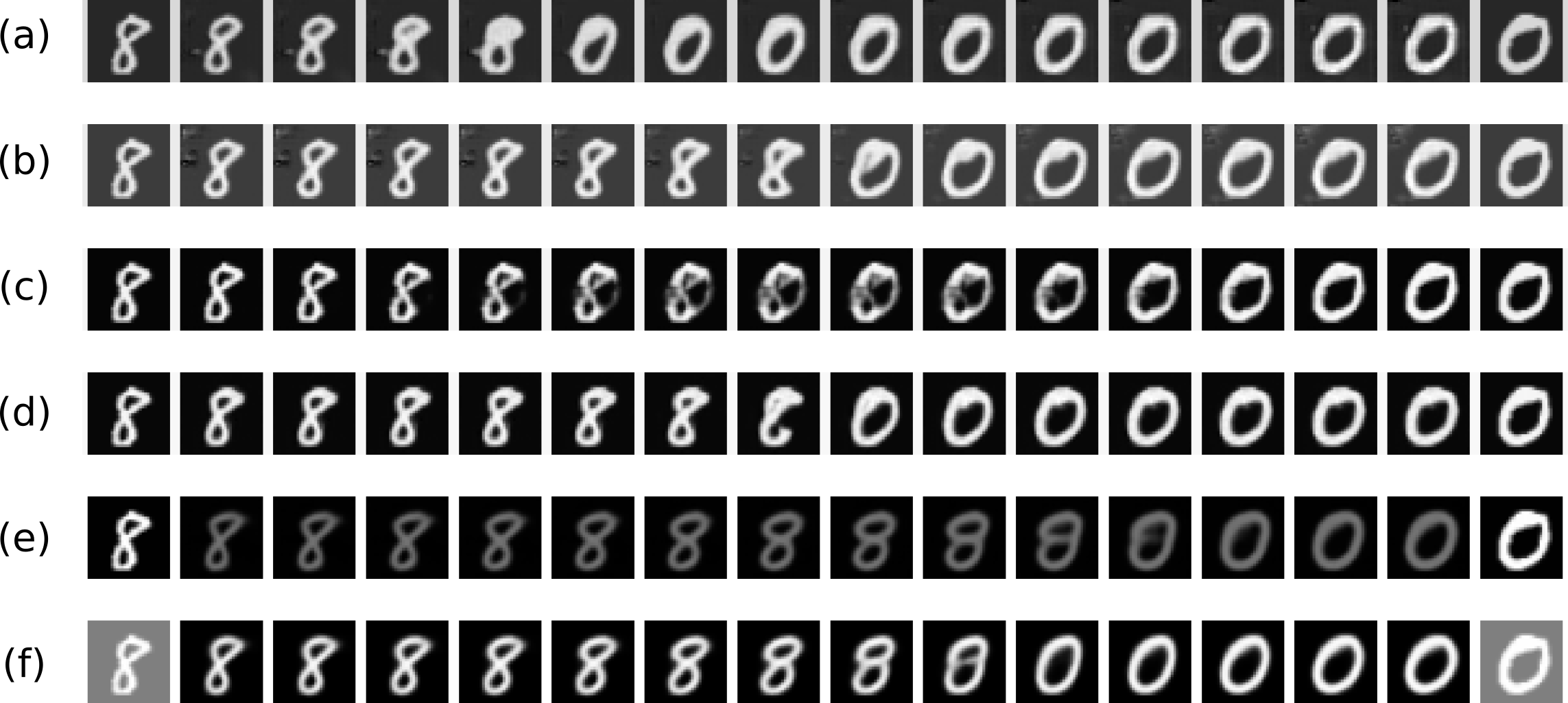}
\caption[MNIST Interpolations]{Interpolations on the MNIST dataset. The panels (a) - (f) show interpolations of the various autoencoders as follows: (a) ACAI, (b) \acronym, (c) IDEC, (d) \acronym+IDEC, (e) VAE, (f) \acronym+VAE.}
\label{fig:interps_mnist}
\end{figure*}

\subsection{Learned Latent Representations}\label{sec:res_learned_representations}
To investigate if the earlier presented intuition can be confirmed in high-dimensional latent spaces, we quantified the shape of the learned embeddings through Principal Component Analysis (PCA). To do this, we trained a baseline autoencoder as well as \acronym on the MNIST dataset. Then, we sub-selected all data points of a given class and encoded the images to receive latent vectors for all images of that class. Subsequently, we performed a PCA on these latent vectors with a self-selected cutoff of 40 in order to obtain the most relevant principal components. After doing this for every class, the components of all classes were normalized and averaged. 

The result of this procedure is shown in figure \ref{fig:illustration_mnist_pca}. In the left panel, the principal components of the baseline model are presented, while the results for the \acronym model are illustrated in the right panel. The first component of the \acronym model explains a visibly smaller amount of variance compared to the first component of the baseline model. Additionally, The differences between the individual components for the \acronym model are in general smaller than for the baseline model. This difference is in line with the intuition outlined in Section \ref{sec:methods} and we will discuss this finding and its implications about the clusters' shapes in Section \ref{sec:discussion}.

\subsection{Clustering Performance}
\begin{figure}[t]%[h]
\centering
\includegraphics[scale=0.52]{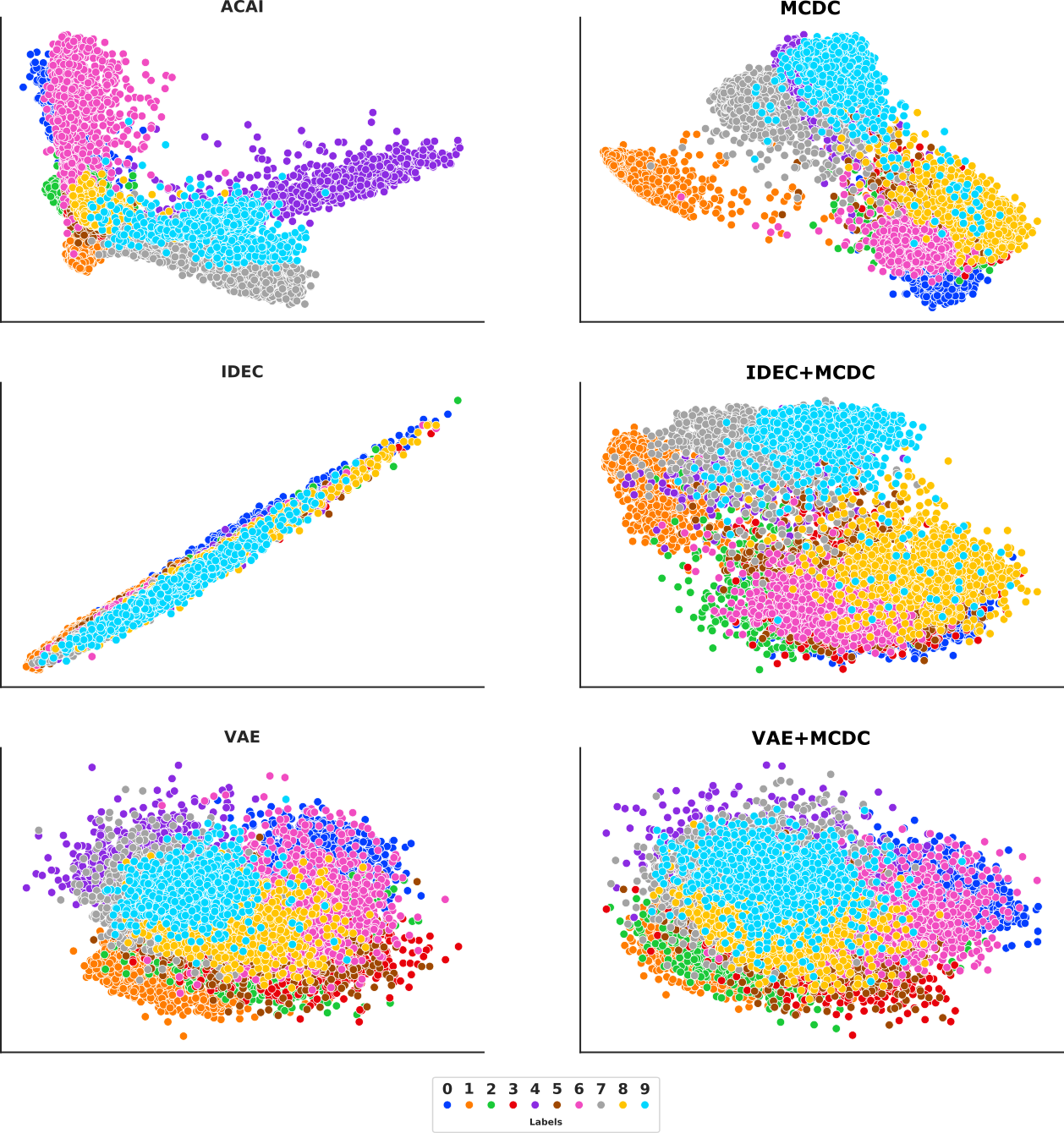}
\caption[pca mnist]{First two PCA components of the learned data point positions in the latent space by the various models on the MNIST dataset.}
\label{fig:pca_mnist}
\end{figure}

The clustering results are shown in table \ref{tab:clusteringResults} and the corresponding PCA embeddings for MNIST are depicted in figure \ref{fig:pca_mnist}. On the MNIST dataset, the combination of the deep clustering model IDEC combined with \acronym training led to the best clustering performance with an accuracy of 97.61\% and an NMI of 93.80\%. On the SVHN as well as CIFAR10, the \acronym model achieved the best performance in terms of accuracy and NMI. Across all datasets, the \acronym model performed better than the ACAI model, suggesting that the mixing consistency loss (see eq. \ref{eq:lossMCDC}) indeed improves clustering performance. Additionally, the models that were trained without and with \acronym (i.e. IDEC and VAE) improved in clustering performance when the \acronym training method was added. This empirical result is present across all datasets, suggesting that adding \acronym to an existing autoencoder model further improves clustering performance of a given autoencoder model.

%% PCA ON MNIST

%-----------------------------------
%	SECTION: Discussion
%-----------------------------------
\section{Discussion}\label{sec:discussion}
As mentioned in Section \ref{sec:res_learned_representations}, the PCA of the \acronym model is characterized by a smaller contribution of the first component and overall smaller differences between the individual components compared to the baseline model. This suggests that the learned data points of each class need more dimensions to be explained well, which can be interpreted as a more spherical shape. This interpretation is supported by the PCA visualization in figure \ref{fig:pca_mnist}. The top right panel of the figure shows the learned positions of the embeddings of the \acronym model projected onto the first two principal components. It becomes apparent that the learned embeddings of the classes appear less elongated as opposed to other models like IDEC and ACAI. 

Furthermore, these interpretations are supported by the accompanying interpolation and clustering results. The interpolations in figure \ref{fig:interps_mnist} illustrate that \acronym based models, especially \acronym and \acronym+IDEC, learn to decode interpolants to look like the first input image for $\alpha \geq 0.5$ while they look like the second input image for $\alpha < 0.5$. This result confirms the effect of adding the mixing consistency loss term as illustrated on the toy dataset in section \ref{sec:methods}. Last but not least, the clustering results suggest that the more spherical shape learned by \acronym may make it easier for the k-Means algorithm to correctly group the data points.

While \acronym improved clustering performance across different models and datasets in our experiments, some questions remain unanswered. For example, we used datasets containing several balanced classes, such that we could uniformly draw samples for interpolations. Future research is needed to investigate whether whether the results obtained in the experiments generalize to datasets with class imbalances or very few classes (e.g. two classes). In these two scenarios, randomly sampling data points for interpolations may lead to substantially more interpolations of the majority classes or, in the case of very few classes, within-class interpolations. Furthermore, recent work in the emerging field of geometric deep learning suggests that the manifold that the latent representations lie on may be curved \cite{arvanitidis2017latent} and, therefore, Euclidean distance may not be the most appropriate distance measure. Instead, it was proposed to calculate the curvature of the manifold and use the geodesic distance as a distance metric. Doing this, deep clustering performance have been reported to drastically increase \cite{shukla2019geometry,yang2018geodesic}. Since the aim of the paper was to introduce \acronym and show its positive impact on adding it to existing methods, combining \acronym with geodesic distance measures was beyond the scope of this paper. Hence, a potential line of research for future work would be to investigate how the \acronym method performs when evaluating the clustering performance using geodesic distance. 

%-----------------------------------
%	SECTION: Conclusion
%-----------------------------------
\section{Conclusion}\label{sec:conclusion}
In the present paper, we proposed \name (\acronym) and applied it to various clustering (non-) specific autoencoder models. We investigated how interpolation and adversarial training influences the learned positions of latent representations in a model's latent space and, eventually, clustering performance. We measured the models’ clustering abilities and found that adding \acronym to a model systematically changed the structure of learned representations of a model, leading to systematic increases in clustering accuracy.

Since the improvement in clustering seems to be robust across autoencoder models, existing autoencoder models can further improve in performance by adding this training method to their training pipeline. In practice, this could lead to better customer segmentation, resulting in individual recommendations based on a person's shopping behaviour. It could also improve the detection of credit card fraud since an anomaly could appear more strongly as an outlier among the normal groups of customer’s behaviour. All in all, adding the \acronym training method to existing deep clustering models could yield a great benefit for people’s lives and businesses.

%-----------------------------------
%	SECTION: References
%-----------------------------------
\bibliography{bibliography.bib}

\begin{thebibliography}{}

\bibitem[\protect\citeauthoryear{Arthur and Vassilvitskii}{2007}]{arthur2007k}
Arthur, D., and Vassilvitskii, S.
\newblock 2007.
\newblock k-means++: The advantages of careful seeding.
\newblock In {\em Proceedings of the eighteenth annual ACM-SIAM symposium on
  Discrete algorithms},  1027--1035.
\newblock Society for Industrial and Applied Mathematics.

\bibitem[\protect\citeauthoryear{Arvanitidis, Hansen, and
  Hauberg}{2018}]{arvanitidis2017latent}
Arvanitidis, G.; Hansen, L.~K.; and Hauberg, S.
\newblock 2018.
\newblock Latent space oddity: on the curvature of deep generative models.
\newblock {\em International Conference on Learning Representations (ICLR)}.

\bibitem[\protect\citeauthoryear{Beckham \bgroup et al\mbox.\egroup
  }{2019}]{beckham2019adversarial}
Beckham, C.; Honari, S.; Verma, V.; Lamb, A.~M.; Ghadiri, F.; Hjelm, R.~D.;
  Bengio, Y.; and Pal, C.
\newblock 2019.
\newblock On adversarial mixup resynthesis.
\newblock In {\em Advances in neural information processing systems},
  4346--4357.

\bibitem[\protect\citeauthoryear{Bengio, Courville, and
  Vincent}{2013}]{bengio2013representation}
Bengio, Y.; Courville, A.; and Vincent, P.
\newblock 2013.
\newblock Representation learning: A review and new perspectives.
\newblock {\em IEEE transactions on pattern analysis and machine intelligence}
  35(8):1798--1828.

\bibitem[\protect\citeauthoryear{Berthelot \bgroup et al\mbox.\egroup
  }{2019a}]{berthelot2019mixmatch}
Berthelot, D.; Carlini, N.; Goodfellow, I.; Papernot, N.; Oliver, A.; and
  Raffel, C.~A.
\newblock 2019a.
\newblock Mixmatch: A holistic approach to semi-supervised learning.
\newblock In {\em Advances in Neural Information Processing Systems},
  5050--5060.

\bibitem[\protect\citeauthoryear{Berthelot \bgroup et al\mbox.\egroup
  }{2019b}]{berthelot2018understanding}
Berthelot, D.; Raffel, C.; Roy, A.; and Goodfellow, I.
\newblock 2019b.
\newblock Understanding and improving interpolation in autoencoders via an
  adversarial regularizer.
\newblock {\em nternational Conference on Learning Representations (ICLR)}.

\bibitem[\protect\citeauthoryear{Goodfellow \bgroup et al\mbox.\egroup
  }{2014}]{goodfellow2014generative}
Goodfellow, I.; Pouget-Abadie, J.; Mirza, M.; Xu, B.; Warde-Farley, D.; Ozair,
  S.; Courville, A.; and Bengio, Y.
\newblock 2014.
\newblock Generative adversarial nets.
\newblock In {\em Advances in neural information processing systems},
  2672--2680.

\bibitem[\protect\citeauthoryear{Guo \bgroup et al\mbox.\egroup
  }{2017}]{guo2017improved}
Guo, X.; Gao, L.; Liu, X.; and Yin, J.
\newblock 2017.
\newblock Improved deep embedded clustering with local structure preservation.
\newblock In {\em International Joint Conference on Artificial Intelligence
  (IJCAI-17)},  1753--1759.

\bibitem[\protect\citeauthoryear{Guo, Mao, and Zhang}{2019}]{guo2019mixup}
Guo, H.; Mao, Y.; and Zhang, R.
\newblock 2019.
\newblock Mixup as locally linear out-of-manifold regularization.
\newblock In {\em Proceedings of the AAAI Conference on Artificial
  Intelligence}, volume~33,  3714--3722.

\bibitem[\protect\citeauthoryear{Hinton and
  Salakhutdinov}{2006}]{hinton2006reducing}
Hinton, G.~E., and Salakhutdinov, R.~R.
\newblock 2006.
\newblock Reducing the dimensionality of data with neural networks.
\newblock {\em Science} 313(5786):504--507.

\bibitem[\protect\citeauthoryear{Jiang \bgroup et al\mbox.\egroup
  }{2017}]{jiang2016variational}
Jiang, Z.; Zheng, Y.; Tan, H.; Tang, B.; and Zhou, H.
\newblock 2017.
\newblock Variational deep embedding: An unsupervised and generative approach
  to clustering.
\newblock  1965--1972.

\bibitem[\protect\citeauthoryear{Kingma and Ba}{2015}]{kingma2014adam}
Kingma, D.~P., and Ba, J.
\newblock 2015.
\newblock Adam: A method for stochastic optimization.
\newblock In {\em International Conference on Learning Representations},
  1965--1978.

\bibitem[\protect\citeauthoryear{Kingma and Welling}{2014}]{kingma2013auto}
Kingma, D.~P., and Welling, M.
\newblock 2014.
\newblock Adam: A method for stochastic optimization.
\newblock In {\em Conference proceedings: papers accepted to the International
  Conference on Learning Representations (ICLR)}.

\bibitem[\protect\citeauthoryear{Krizhevsky and
  Hinton}{2009}]{krizhevsky2009learning}
Krizhevsky, A., and Hinton, G.
\newblock 2009.
\newblock Learning multiple layers of features from tiny images.
\newblock Technical Report~4, University of Toronto.

\bibitem[\protect\citeauthoryear{Kuhn}{2005}]{kuhn2005hungarian}
Kuhn, H.~W.
\newblock 2005.
\newblock The hungarian method for the assignment problem.
\newblock {\em Naval Research Logistics (NRL)} 52(1):7--21.

\bibitem[\protect\citeauthoryear{LeCun, Bengio, and
  Hinton}{2015}]{lecun2015deep}
LeCun, Y.; Bengio, Y.; and Hinton, G.
\newblock 2015.
\newblock Deep learning.
\newblock {\em Nature} 521(7553):436.

\bibitem[\protect\citeauthoryear{LeCun}{1998}]{lecun1998mnist}
LeCun, Y.
\newblock 1998.
\newblock The mnist database of handwritten digits.
\newblock {\em http://yann. lecun. com/exdb/mnist/}.

\bibitem[\protect\citeauthoryear{Maas, Hannun, and
  Ng}{2013}]{maas2013rectifier}
Maas, A.~L.; Hannun, A.~Y.; and Ng, A.~Y.
\newblock 2013.
\newblock Rectifier nonlinearities improve neural network acoustic models.
\newblock In {\em Proc. icml}, number~1, ~3.

\bibitem[\protect\citeauthoryear{Min \bgroup et al\mbox.\egroup
  }{2018}]{min2018survey}
Min, E.; Guo, X.; Liu, Q.; Zhang, G.; Cui, J.; and Long, J.
\newblock 2018.
\newblock A survey of clustering with deep learning: From the perspective of
  network architecture.
\newblock {\em IEEE Access} 6:39501--39514.

\bibitem[\protect\citeauthoryear{Netzer \bgroup et al\mbox.\egroup
  }{2011}]{netzer2011reading}
Netzer, Y.; Wang, T.; Coates, A.; Bissacco, A.; Wu, B.; and Ng, A.~Y.
\newblock 2011.
\newblock Reading digits in natural images with unsupervised feature learning.

\bibitem[\protect\citeauthoryear{Ng, Jordan, and Weiss}{2002}]{ng2002spectral}
Ng, A.~Y.; Jordan, M.~I.; and Weiss, Y.
\newblock 2002.
\newblock On spectral clustering: Analysis and an algorithm.
\newblock In {\em Advances in neural information processing systems},
  849--856.

\bibitem[\protect\citeauthoryear{Odena, Dumoulin, and
  Olah}{2016}]{odena2016deconvolution}
Odena, A.; Dumoulin, V.; and Olah, C.
\newblock 2016.
\newblock Deconvolution and checkerboard artifacts.
\newblock {\em Distill} 1(10):e3.

\bibitem[\protect\citeauthoryear{Paparrizos and
  Gravano}{2015}]{paparrizos2015k}
Paparrizos, J., and Gravano, L.
\newblock 2015.
\newblock k-shape: Efficient and accurate clustering of time series.
\newblock In {\em Proceedings of the 2015 ACM SIGMOD International Conference
  on Management of Data},  1855--1870.
\newblock ACM.

\bibitem[\protect\citeauthoryear{Rokach and
  Maimon}{2005}]{rokach2005clustering}
Rokach, L., and Maimon, O.
\newblock 2005.
\newblock Clustering methods.
\newblock In {\em Data mining and knowledge discovery handbook}. Springer.
\newblock  321--352.

\bibitem[\protect\citeauthoryear{Schmidhuber}{2015}]{schmidhuber2015deep}
Schmidhuber, J.
\newblock 2015.
\newblock Deep learning in neural networks: An overview.
\newblock {\em Neural networks} 61:85--117.

\bibitem[\protect\citeauthoryear{Shukla \bgroup et al\mbox.\egroup
  }{2019}]{shukla2019geometry}
Shukla, A.; Uppal, S.; Bhagat, S.; Anand, S.; and Turaga, P.
\newblock 2019.
\newblock Geometry of deep generative models for disentangled representations.
\newblock {\em arXiv preprint arXiv:1902.06964}.

\bibitem[\protect\citeauthoryear{Van Den~Oord, Vinyals, and
  others}{2017}]{van2017neural}
Van Den~Oord, A.; Vinyals, O.; et~al.
\newblock 2017.
\newblock Neural discrete representation learning.
\newblock In {\em Advances in Neural Information Processing Systems},
  6306--6315.

\bibitem[\protect\citeauthoryear{Verma \bgroup et al\mbox.\egroup
  }{2018}]{verma2018manifold}
Verma, V.; Lamb, A.; Beckham, C.; Courville, A.; Mitliagkis, I.; and Bengio, Y.
\newblock 2018.
\newblock Manifold mixup: Encouraging meaningful on-manifold interpolation as a
  regularizer.
\newblock {\em arXiv preprint arXiv:1806.05236}.

\bibitem[\protect\citeauthoryear{Verma \bgroup et al\mbox.\egroup
  }{2019}]{verma2019interpolation}
Verma, V.; Lamb, A.; Kannala, J.; Bengio, Y.; and Lopez-Paz, D.
\newblock 2019.
\newblock Interpolation consistency training for semi-supervised learning.
\newblock {\em arXiv preprint arXiv:1903.03825}.

\bibitem[\protect\citeauthoryear{Wang, Cui, and Zhu}{2016}]{wang2016structural}
Wang, D.; Cui, P.; and Zhu, W.
\newblock 2016.
\newblock Structural deep network embedding.
\newblock In {\em Proceedings of the 22nd ACM SIGKDD international conference
  on Knowledge discovery and data mining},  1225--1234.

\bibitem[\protect\citeauthoryear{Wold, Esbensen, and
  Geladi}{1987}]{wold1987principal}
Wold, S.; Esbensen, K.; and Geladi, P.
\newblock 1987.
\newblock Principal component analysis.
\newblock {\em Chemometrics and intelligent laboratory systems} 2(1-3):37--52.

\bibitem[\protect\citeauthoryear{Xie, Girshick, and
  Farhadi}{2016}]{xie2016unsupervised}
Xie, J.; Girshick, R.; and Farhadi, A.
\newblock 2016.
\newblock Unsupervised deep embedding for clustering analysis.
\newblock In {\em International conference on machine learning},  478--487.

\bibitem[\protect\citeauthoryear{Yang \bgroup et al\mbox.\egroup
  }{2018}]{yang2018geodesic}
Yang, T.; Arvanitidis, G.; Fu, D.; Li, X.; and Hauberg, S.
\newblock 2018.
\newblock Geodesic clustering in deep generative models.
\newblock {\em arXiv preprint arXiv:1809.04747}.

\end{thebibliography}
\bibliographystyle{aaai}
\end{document}